\PassOptionsToPackage{table,dvipsnames}{xcolor}
\documentclass[sigconf,nonacm]{acmart} 

\usepackage{tcolorbox}

\graphicspath{{./images/}} 

\usepackage{graphicx}
\usepackage{makecell}
\usepackage{enumitem}

\usepackage{booktabs}
\usepackage[table]{xcolor}
\usepackage{tcolorbox}
\usepackage{array}
\usepackage[dvipsnames]{xcolor}

\begin{document}

\title[PaPaformer]{PaPaformer: Language Model from Pre-trained Parallel Paths}

\author{Joonas Tapaninaho}
\affiliation{%
  \institution{University of Oulu, Faculty of Information Technology and Electrical Engineering, CMVS}
  \city{Oulu}
  \country{Finland}
}

\author{Mourad Oussalah}
\affiliation{%
  \institution{University of Oulu, Faculty of Information Technology and Electrical Engineering, CMVS}
  \city{Oulu}
  \country{Finland}
}

\renewcommand{\shortauthors}{Tapaninaho}

\begin{abstract}
The training of modern large-language models requires an increasingly amount of computation power and time. Even smaller variants, such as small-language models (SLMs), take several days to train in the best-case scenarios, often requiring multiple GPUs. This paper explores methods to train and evaluate decoder-only transformer-based language models in hours instead of days/weeks. We introduces \textit{PaPaformer}, a decoder-only transformer architecture variant, whose lower-dimensional parallel paths are combined into larger model. The paper shows that these lower-dimensional paths can be trained individually with different types of training data and then combined into one larger model. This method gives the option to reduce the total number of model parameters and the training time with increasing performance. Moreover, the use of parallel path structure opens interesting possibilities to customize paths to accommodate specific task requirements.
\end{abstract}

\maketitle

\newcommand{\myheading}[1]{\paragraph*{#1.}}





\section{Introduction}
Large Language Models (LLMs) have significantly transformed people's daily lives across numerous domains, following the revolutionary introduction of the autoregressive transformer \cite{vaswani2023attentionneed} and its adaptation to the decoder-only transformer, known as the Generative Pretrained Transformer (GPT) \cite{Radford2018ImprovingLU}. For years, state-of-the-art (SOTA) LLMs such as GPT-3 \cite{brown2020language}, LLaMA \cite{touvron2023llama}, and Falcon \cite{penedo2023falcon} have relied on dense architectures, where every parameter in the model contributes to the prediction of the next token. However, these dense models have recently encountered a compelling alternative with the emergence of DeepSeek-V2 \cite{deepseek2024deepseekv2}, which employs a Mixture-of-Experts (MoE) mechanism to activate only a subset of model parameters during inference, achieving computational efficiency without compromising performance reduction. 

Despite recent progress, most MoE approaches operate strictly within the feedforward sublayers and rely on dynamically selected routing. This leaves open the question of how well structured, modular parallelism might perform within the core architecture itself when combined with MoE-style routing.

This work investigates an alternative architectural approach by introducing fixed parallel processing paths into the decoder-only Transformer structure and combining them with a lightweight routing mechanism. It also hypothesizes that such a design can enable parameter-efficient learning while supporting modular reuse in cases where the parallel branches are pretrained independently on different data domains and later composed into a combined model.

To further explore this idea, a compact decoder-only transformer architecture is proposed. In this architecture, two main layers encapsulate lower-dimensional parallel paths, reducing the total number of model parameters compared to a fully stacked architecture. In addition, the parallel paths are trained separately on different domains (narrative content and math/instructional prompts) and are later composed into a combined model through continued pre-training.

This work focuses on evaluating whether the resulting combined model can not only improve the performance as compared to large dense baseline models, but also learn to selectively route inputs to the appropriate parallel subpath based on content type.

Due to limited computational resources, all experiments in this work were conducted using small-scale models (under 30M parameters) trained either solely on TinyStory datasets or on a combined dataset that includes a small subset of OpenMathInstruct-1. These models enable fast architectural iteration and, despite their small size, offer valuable insights into routing dynamics and path specialization. More specifically, the main contributions of this work are summarized below.

\begin{enumerate}
\item We showed that with an appropriate connection mechanism, parallel model variants can outperform or match the performances of dense baselines at comparable scale.

\item We showed that routing behavior often struggles to fully and consistently utilize domain-specific pretrained paths.

\item We showed that evenly distributed path selection and utilization tend to correlate with lower performance, whereas models that favor a more dominant path often achieve better task alignment and overall results.

\item We unveiled that modular parallel design offers a novel perspective on architectural sparsity and weight reuse, showing the pathway toward more efficient and composable language models.
\end{enumerate}

\section{Related Works}
\subsection{Decoder-only Language Models}
The Transformer architecture introduced by Vaswani \textit{et al.} \cite{vaswani2023attentionneed} forms the foundation of all modern language modeling. Its decoder-only variant was popularized by the GPT framework \cite{Radford2018ImprovingLU}, and became standard for autoregressive text generation. Large-scale models such as GPT-3, LLaMA, and Falcon follow this design, which relies on dense architectures where all model parameters are used during every forward pass. Although this design achieves strong performance, it also incurs significant computational overhead and also limited modularity.

\subsection{Sparse Activation and Mixture of Experts (MoE)}
 
To reduce the computational cost of dense models, several works have explored more sparse activation techniques, where only a subset of parameters is used during forward pass. Mixture-of-Experts architectures such as GShard \cite{lepikhin2020gshardscalinggiantmodels}, Switch Transformer \cite{fedus2022switchtransformersscalingtrillion}, GLaM~\cite{du2022glam} and DeepSeek-V2 \cite{deepseek2024deepseekv2} flexibly activate only a smaller number of experts depending on input tokens. These models achieve high performance while reducing the number of active parameters in each forward pass. The core insight is that not all parameters of such models need to participate in every computation, which opens the door to more efficient and scalable model designs. Our work is built on this idea, introducing parallel paths that are selectively routed in a similar style, but implemented with a fixed structure rather than dynamic expert selection.

\subsection{Parallel Architectures}
Although popular LLMs primarily use a sequential pipeline to process input, some models have explored parallel designs within the Transformer architecture. For example, models such as PaLM~\cite{chowdhery2022palm} and Branchformer~\cite{peng2022branchformerparallelmlpattentionarchitectures} incorporate parallelism within feedforward and attention mechanisms to enhance representational capacity.
Additionally, certain Transformer variants employ parallel attention mechanisms to increase model expressiveness. However, to date, no popular work has yet combined layer-level parallelism with sparse activation to achieve both parameter efficiency and architectural flexibility in decoder-only language models.
This work introduces an alternative design in which parallel paths can be built from smaller pretrained models and combined into part of unified larger architecture, offering a path towards more flexible compositional modular language models.

\subsection{Tiny Language Models}
Some notable work has shown interest in reliably evaluating small-scale language models, which require only limited computational resources. Datasets like TinyStories \cite{eldan2023tinystoriessmalllanguagemodels} and benchmarks adapted from small models help assess generalization, syntax understanding, and reasoning capabilities in models that contain fewer than 100 million parameters. Active competitions such as BaByLM \cite{babylm-2024} also emphasize that architectural innovations can be meaningfully evaluated on the scale of the SLM and even Tiny-Language Model (TLM). These efforts support the use of compact and \textit{quickly trainable} models with small training dataset to test architectural modifications. This idea is implemented in our work, where parallel-path architectures with sparse routing and without requiring full-scale pretraining are tested using a small training dataset.

\section{Methodology}

\subsection{Overview of the Base Architecture}
The proposed and base models are built upon the decoder-only transformer architecture introduced in LLaMA~\cite{touvron2023llama}, which is an efficient foundation widely adopted in open large-language models due to its open informational availability and competitive performances.

\begin{figure}[ht]
\centering
\includegraphics[width=0.35\textwidth]{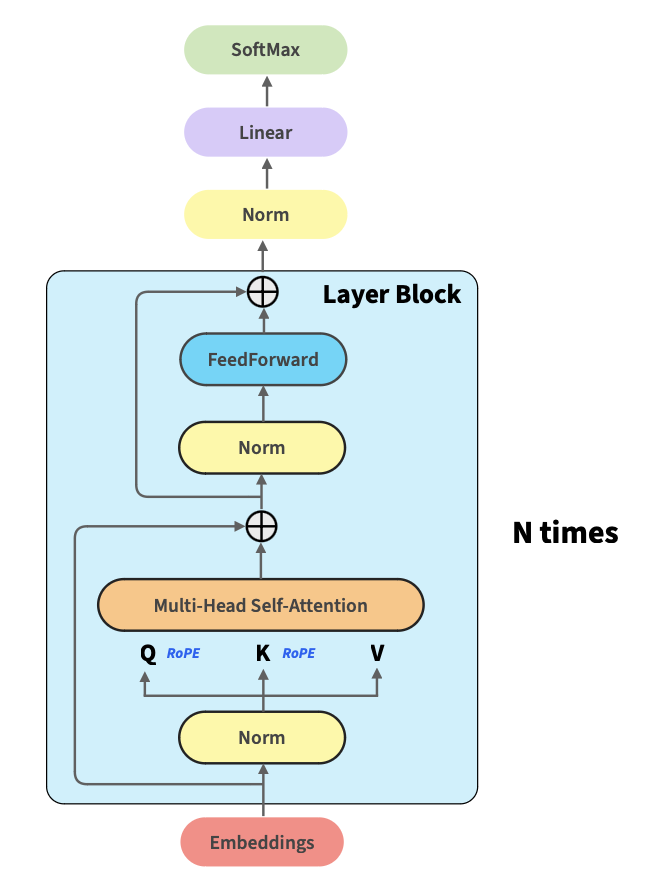}
\Description{..}
\caption[]{Base architecture used as the foundation in this work. The diagram shows a single \textit{Layer Block}, which is stacked $N$ times to form the full model. This structure is representative of the LLaMA architecture; \textit{Norm} corresponds to \textit{RMSNorm}, and \textit{FeedForward} corresponds to \textit{SwiGLU}.}
\label{fig:base-architecture}
\end{figure}

The LLaMA architecture follows the standard Transformer decoder stack visualized in Figure~\ref{fig:base-architecture}, consisting of a multi-head self-attention mechanism or some variant of it, position-wise feedforward networks (FFNs), normalization, and residual connections. Key enhancements compared to the original GPT-style architecture include the use of rotary positional embeddings (RoPE) to encode input token positions, and the application of additional normalization before the attention and FFN sub-layers to improve training stability.

Each layer block processes the input sequence through a self-attention module followed by a feedforward module, with residual connections surrounding both subcomponents. The model is trained using an unsupervised autoregressive next-token prediction given all previous tokens in the sequence. This architecture has demonstrated strong performance across a range of benchmarks and has also inspired other architecture variants as well.

In this work, the core structure of LLaMA is retained while modifying internal components to incorporate parallel processing paths as described in the following subsections. The intention is to test whether structured architectural parallelism can improve model representational capacity and parameter efficiency without requiring large-scale models or as many parameters.

\subsection{Parallel Path Integration}
\label{sec:parallel-path-integration}
To investigate the impact and advantage of architectural parallelism in decoder-only language models, this work proposes a modified design that varies between standard full-size layer blocks and intermediate modules, including \textit{parallel processing paths}. This hybrid structure is built following the base LLaMA architecture, while introducing high modularity and flexible parallel path composition.

\begin{figure}[ht]
\centering
\includegraphics[width=0.35\textwidth]{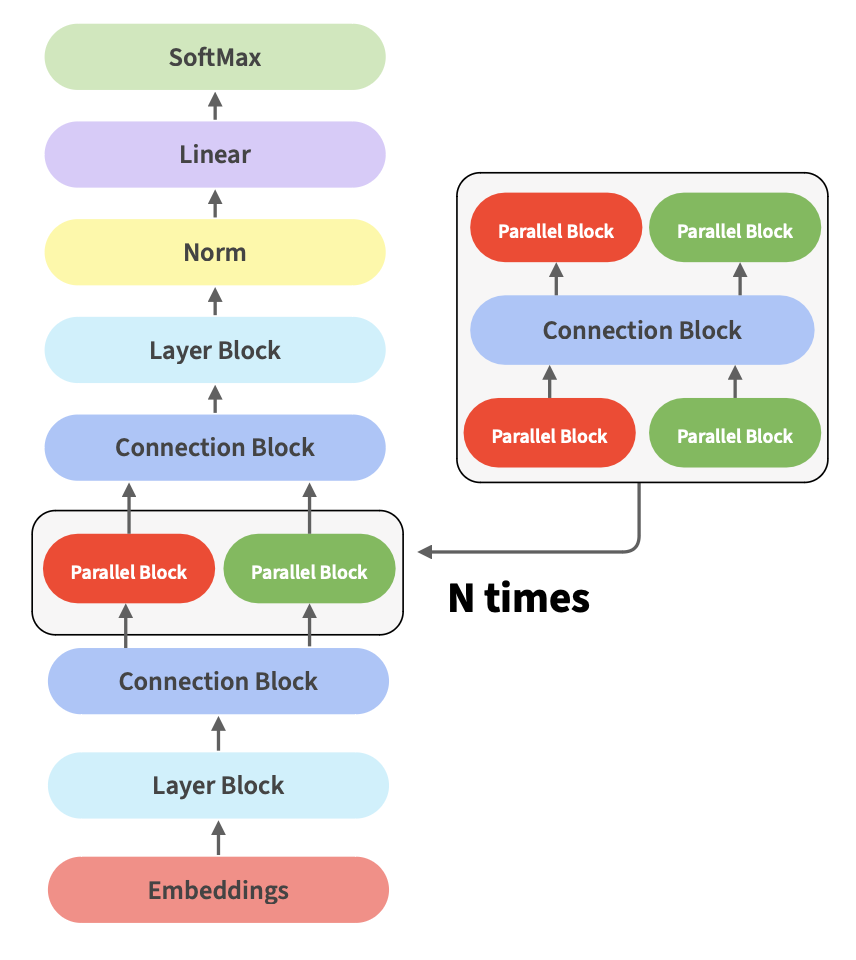}
\Description{..}
\caption[]{Parallel-path architecture used in this work. The model alternates between standard \textit{Layer Blocks} and stacks of modular \textit{Parallel Blocks} in parallel layers. \textit{Connection Blocks} are included before and after each parallel layer to exchange information and (if needed) modify dimensionality. Multiple Parallel Block layers can be stacked ($N$ times) before second \textit{Layer Block}.}
\label{fig:parallel-architecture}
\end{figure}

As illustrated in Figure~\ref{fig:parallel-architecture}, at the beginning, we used a full-size \textit{Layer Block}, followed by a \textit{Connection Block} whose role is to optionally adjust the dimensionality and serve as a communication bridge between parallel paths. The central part of the model consists of multiple stacked \textit{Parallel Layers}, where each block contains two or more independent sub-paths.

Between the first-layer block and each parallel layer, \textit{Connection Block} enables cross-path information flow and transformation. Importantly, Connection Blocks are used not only to modify dimensionality but also to promote integration of information between the specialized sub-paths.

More formally, let $x \in \mathbb{R}^{d}$ represent the input for a parallel layer, and let $\{f_1, f_2, \dots, f_k\}$ represent $k$ parallel blocks, each operating at dimension $d'$. Parallel blocks produce outputs $f_i(x) \in \mathbb{R}^{d'}$, which are concatenated to form:

\begin{equation}
y = \text{Concatenate}(f_1(x), \dots, f_k(x)) \in \mathbb{R}^{k \cdot d'}
\end{equation} \\
In this work, every parallel variant is such that $d' = d / k$, so that the concatenated output naturally returns to the original layer block hidden with dimension $d$, avoiding the need for projection. However, a Connection Block can be applied to enable richer interaction across parallel paths and to support structural variants, where, for example, not all paths use the same dimension $d'$.

This approach allows the model to incorporate modular and specialized sub-paths while preserving communication between them and maintaining compatibility with standard decoder-only Transformer training regimes.

\subsection{Parallel Path Variants}
This work implements and evaluates two different architectural strategies to connect and collate the outputs of the parallel paths introduced in Section~\ref{sec:parallel-path-integration}, the \textit{Share Linear} variant and the \textit{Gumbel MoE} variants. Both approaches preserve the same high-level structure but differ in the way the parallel paths are connected and routed.

\subsubsection{Share Linear}
\label{sec:share-linear}
\begin{figure}[ht]
\centering
\includegraphics[width=0.2\textwidth]{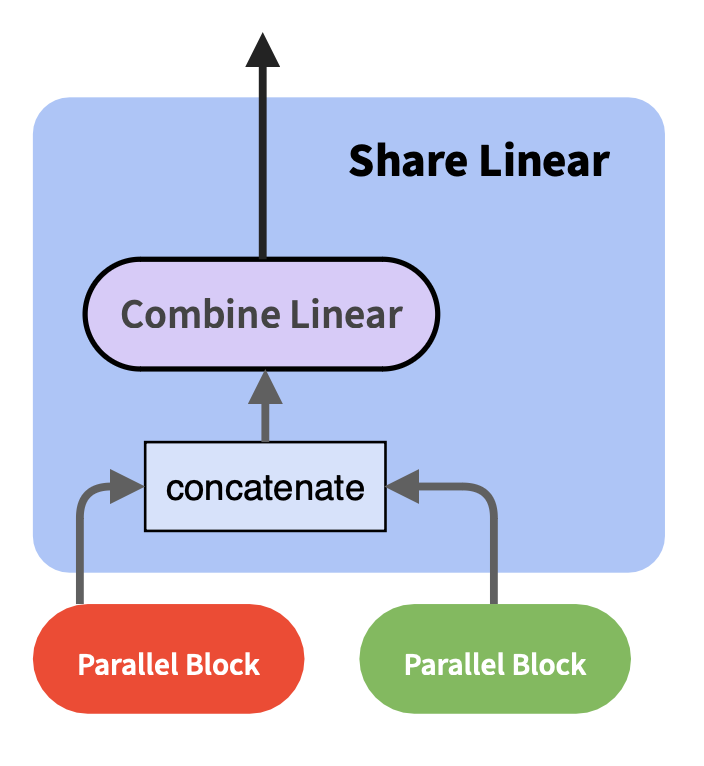}
\Description{..}
\caption[]{
Share Linear version of Connection Block}
\label{fig:ShareLinear}
\end{figure}

The \textit{Share Linear} variant uses a simple linear strategy to connect and compile information across parallel paths as shown in Figure~\ref{fig:ShareLinear}. Specifically, after the initial \textit{Layer Block}, the output is passed through a \textit{Connection Block} implemented as a linear projection that reduces the hidden dimension from $d_{\text{Layer\_Block}}$ to $d_{\text{Parallel\_Block}}$. This lower-dimensional Connection Block output is then fed into a parallel layer, which contains $k$ \textit{Parallel Blocks}, each operating independently with the same structure. After each parallel layer, the Connection Block (share linear) is applied. 

This architecture variant supports efficient forward computation, where weight reuse (parallel paths were initialized from pre-trained models), enables independent training of each parallel path before combination, while maintaining full compatibility with standard decoder-only Transformer training workflows.

\subsubsection{Gumbel MoE}
The \textit{Gumbel MoE} variants use a straightforward mixture-of-experts (MoE) routing method inside the \textit{Connection Block} to dynamically learn how to combine outputs from parallel paths. Unlike the traditional MoE setup where a router is used to select k-experts (FeedForward Neural Networks) among N possible experts, in this design variant, the "parallel paths" play the role of the experts and the router selects a given "parallel path" or a combination of them which substrantially reduces the number of parameters as compared to traditional MoE.
 As in the \textit{Share Linear} variant (Section~\ref{sec:share-linear}), the output from the initial \textbf{Layer Block} is passed through a \textbf{Connection Block} implemented as a linear projection that reduces the hidden dimension from $d_{\text{Layer\_Block}}$ to $d_{\text{Parallel\_Block}}$.

\begin{figure}[ht]
\centering
\includegraphics[width=0.5\textwidth]{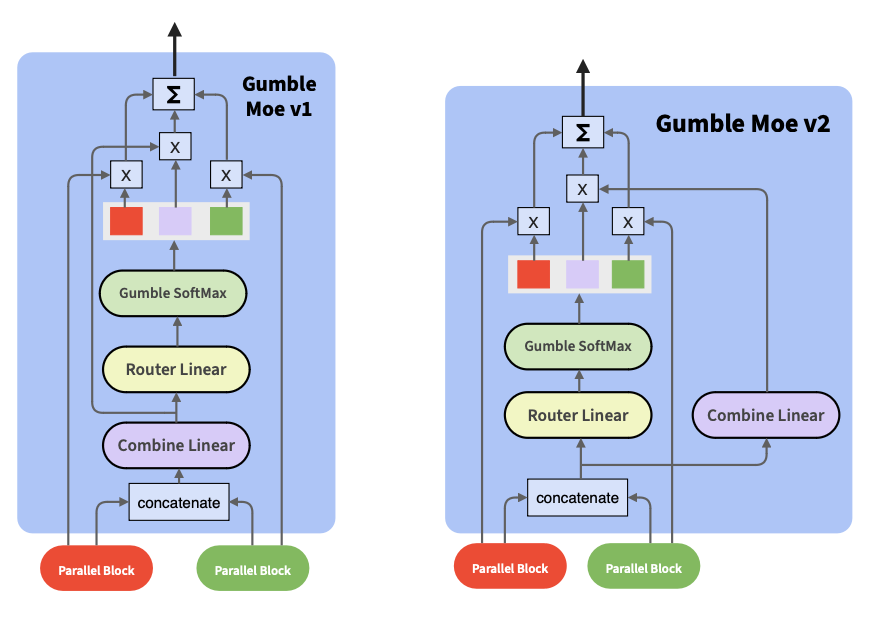}
\Description{..}
\caption[]{
Comparison of two Gumbel MoE routing strategies for collating outputs from parallel blocks. 
\texttt{Left} (Gumbel MoE v1): The outputs from all parallel blocks are first concatenated and then passed through a Combine Linear layer. The resulting combined output is fed into a Router Linear layer, whose output scores are converted by the Gumbel-Softmax into weights. These weights are then used to compute a weighted sum of the original path outputs and their combined representation. \texttt{Right} (Gumbel MoE v2): In contrast to Gumbel MoE v1, the parallel outputs are concatenated and scored directly by the Router Linear, while the combined representation is computed separately using a dedicated Combine Linear layer. Both strategies maintain differentiability and support soft expert-style selection in a lightweight form suitable for compact models.
}
\label{fig:gumbel-moe-variants}
\end{figure}

In contrast to using only a linear combination after each parallel layer, the Gumbel MoE variant combines the outputs of the parallel blocks using the Gumbel-Softmax routing as shown in Figure~\ref{fig:gumbel-moe-variants}. 

The purpose of applying the Gumbel-Softmax trick~\cite{jang2017categorical} is to allow differentiable selection among parallel paths, enabling backpropagation, while maintaining discrete path selection behavior during training. Specifically, we explore two routing configurations:

\begin{itemize}
    \item \textbf{Variant 1 – Gumbel routing from combined representation:} The outputs from the parallel blocks are first concatenated and passed through a \textit{Combine Linear} layer:
    \begin{equation}
    x_{\text{comb}} = W_{\text{Combine}}([f_1(x), \dots, f_k(x)])
    \end{equation}
    
The combined result is then passed to the \textit{Router Linear} layer, before the \textit{Gumbel-Softmax} gate computes weights over the individual block outputs, which are subsequently summed as follows:

    \begin{equation}
    y = \sum_{i=1}^k \pi_i \cdot f_i(x) + \pi_{\text{comb}} \cdot x_{\text{comb}}
    \end{equation}

where $\boldsymbol{\pi} = [\pi_1, \dots, \pi_k, \pi_{\text{comb}}]$ are the routing weights produced via Gumbel-Softmax:
     \[ \text{Gumbel\_Softmax}(x_{\text{comb}})\]

    \item \textbf{Variant 2 – Routing from concatenated outputs:} The concatenated outputs are first passed through a \textit{Router Linear} layer to produce routing logits:

        \begin{equation}
        x_{\text{router}} = W_{\text{Router}}([f_1(x), \dots, f_k(x)]) 
        \end{equation}
        \begin{equation}
        \boldsymbol{\pi} = \text{Gumbel\_Softmax}(x_{\text{router}})
        \end{equation}
    
    Simultaneously, a combined representation is computed using a separate \textit{Combine Linear} layer:
    
    \begin{equation}
    x_{\text{comb}} = W_{\text{Combine}}([f_1(x), \dots, f_k(x)])
    \end{equation}
    
    The final output is then obtained as:
    
    \begin{equation}
    y = \sum_{i=1}^k \pi_i \cdot f_i(x) + \pi_{\text{comb}} \cdot x_{\text{comb}}
    \end{equation}
    
\end{itemize}

Table~\ref{tab:gumbel-moe-symbolic-dims} shows the dimensionality differences of those variants in the corresponding linear components.
\vspace{2ex}

\begin{table}[ht]
\centering
\caption[]{Input and output dimensions (resp. dim\_in and dim\_out) of linear layers in Gumbel MoE v1 and v2, where $k$ is the number of parallel paths and $d'$ is the hidden size of each path.}
\begin{tabular}{|l|c|c|}
\hline
\textbf{Layer} & \textbf{dim\_in} & \textbf{dim\_out} \\
\hline
Gumbel MoE v1 - Combine Linear & $k \cdot d'$ & $d'$ \\
\hline
Gumbel MoE v1 - Router Linear  & $d'$         & $k{+}1$ \\
\hline
Gumbel MoE v2 - Router Linear  & $k \cdot d'$ & $k{+}1$ \\
\hline
Gumbel MoE v2 - Combine Linear & $k \cdot d'$ & $d'$ \\
\hline
\end{tabular}
\vspace{0.5em}
\label{tab:gumbel-moe-symbolic-dims}
\end{table}
\vspace{-1.75em}

\noindent\textbf{Auxiliary Losses.}  
To encourage meaningful routing behavior, two simple auxiliary regularization terms are applied during the pretraining phase: \textit{entropy regularization} and \textit{load balancing}.

As motivation, the purpose of these terms is to jointly promote a more stable parallel-path training and improved generalization, particularly under low-resource or small-model settings.

This dual regularization strategy uses ideas from prior mixture-of-experts architectures~\cite{shazeer2017outrageously, fedus2021switch, du2022glam} employing routing stability and expert load balancing. However, a slight formulating was introduced, concerning the way our model combines token-level entropy regularization with batch-level load balancing to achieve stabilize training and improve generalization in parallel path models architecture.
\noindent More specifically, the entropy loss encourages the router to maintain soft probability distributions across parallel experts for each token. This is important for preventing premature convergence to a single path and supports continued exploration in routing decisions. Formally, the entropy loss is written as:
\vspace{-0.25em}
\begin{equation}
\mathcal{L}_{\text{entropy}} = -\frac{1}{N} \sum_{b=1}^B \sum_{t=1}^T \sum_{i=1}^k \pi_{b,t,i} \log(\pi_{b,t,i})
\end{equation}
where $\pi_{b,t,i}$ is the routing probability for expert $i$ at batch $b$, token position $t$, and $N = B \cdot T$ is the total number of tokens.

The purpose of the load-balancing loss is to mitigate the global expert imbalance by encouraging equal utilization of all experts across the batch:
\vspace{-0.25em}

\begin{equation}
\bar{\pi}_i = \frac{1}{N} \sum_{b=1}^B \sum_{t=1}^T \pi_{b,t,i}, \quad 
\mathcal{L}_{\text{load}} = - \sum_{i=1}^k \bar{\pi}_i \log(\bar{\pi}_i)
\end{equation}

Where, $\bar{\pi}_i$ represents the average routing probability of expert $i$ across all tokens in the batch.

Therefore, the final loss function becomes:
\begin{equation}
\mathcal{L}_{\text{total}} = \mathcal{L}_{\text{CE}} + \lambda_{\text{entropy}} \cdot \mathcal{L}_{\text{entropy}} + \lambda_{\text{load}} \cdot \mathcal{L}_{\text{load}}
\end{equation}
where $\mathcal{L}_{\text{CE}}$ is the standard cross-entropy loss, and \\ 
    $\lambda_{\text{entropy}} = \lambda_{\text{load}} = 0.01$.

\subsection{Training Setup and Implementation Details}
This section describes the training setup used to pre-train the proposed architecture variants and comparison models, including model and training configurations, optimization strategies, and hardware considerations. Although the implementation is restricted to a compact model size due to computational constraints, the training procedure in other ways closely follows standard practices commonly used in autoregressive language modeling.

\subsubsection{Model Configuration.}
In our work, the models that follow the base LLaMA architecture~\cite{touvron2023llama} consist only of stacked \textit{Layer Blocks} with hidden dimensions of 128, 192, or 256. On the other hand, the models that use the parallel architecture described in Sections~3.2--3.3 include an initial \textit{Layer Block}, followed by a \textit{Connection Block}, then stacked parallel layers, and finally a concluding \textit{Layer Block}, where each \textit{Parallel Layer} is composed of two \textbf{Parallel Blocks} followed by a \textit{Connection Block}. In the case of the parallel architecture, the \textit{Layer Blocks} use a hidden dimension of 256, which is projected down to 128 using a \textit{Connection Block} implemented as a linear layer, in order to match the hidden dimensions (128) used by the \textit{Parallel Blocks}.

In the first and second training phases, all models have a total of 8 layers (blocks), where the parallel architectures used 3 \textit{Layer Blocks} and 2 \textit{Parallel Layers}. Table~\ref{tab:architecture-comparison} summarizes the key architectural hyperparameters used across three configurations: the Base models, the shared configuration for Path\_1 and Path\_2, and the final Parallel architecture variants.

\begin{table}[ht]
\centering
\caption[]{Comparison of architectural hyperparameters across the Base model, the shared Path\_1 \& Path\_2 configuration, and the parallel model variants.}
\begin{tabular}{lccc}
\toprule
\textbf{Hyperparameter} & \makecell{\textbf{Base} \\ \textbf{Model}} & \makecell{\textbf{Path} \\  \textbf{1 \& 2}}

& \makecell{\textbf{Parallel Model}\\\textbf{(Layer + Parallel)}} \\
\midrule
Vocabulary size     & $\sim$ 50K    & $\sim$ 50K  & $\sim$ 50K  \\
\midrule
$d_{\text{model}}$  & \makecell{256, \\ or 192}      & 128       & \makecell{256 (Layer) \\ 128 (Parallel)} \\
\midrule
Layers (Blocks) & 8         & 3         & \makecell{ 2 (Layer) \\  3 (Parallel)}   \\
\midrule
Parallel blocks             & -         & -         & 2 \\
\midrule
Heads               & \makecell{8 \\ or 6}        & 4         & \makecell{8 (Layer) \\ 4 (Parallel) }\\
\midrule
Head dimension                & 32        & 32        & 32 (both) \\
\midrule
FF hidden size       & \makecell{1024 \\ or 728}     & 512       & \makecell{1024 (Layer) \\ 
512 (Path) }\\
\midrule
FF type       & \makecell{SwiGLU}      & SwiGLU      & SwiGLU (both) \\
\midrule
Normalization       & \makecell{RMS}      & RMS      & RMS (both) \\
\midrule
Attention       & MHA       & MHA       & MHA (both) \\
\midrule
Max seq len       & 256       & 256       & 256  \\
\midrule
\makecell{Total Parameter \\ (millions)}      & \makecell{32M \\ or 22.5M}     & 13.5M       & 28.5M \\
\bottomrule
\end{tabular}
\vspace{0.5em}
\label{tab:architecture-comparison}
\end{table}

\textbf{Training Objective and Optimization:}
All models were trained using the \textit{causal language modeling objective}, which minimizes the negative log-likelihood of the next token given the previous tokens. The loss function is the standard \textit{cross-entropy loss} computed over the vocabulary.

This work adopts the optimization settings shown in Table~\ref{tab:pretrain-hyperparams}. The dropout is optionally applied during Parallel Blocks training to improve generalization.

\begin{table}[ht]
\centering
\caption[]{Pre-training hyperparameters used across all experiments.}
\begin{tabular}{ll}
\toprule
\textbf{Hyperparameter}     & \textbf{Value} \\
\midrule
Optimizer                   & AdamW \\
Learning rate               & $5 \times 10^{-4}$ \\
Batch size                  & 32 \\
Training epochs             & 2 \\
Gradient accumulation steps & 8 \\
Weight decay                & 0.1 \\
Adam betas                  & (0.9, 0.95) \\
Adam epsilon                & $1 \times 10^{-5}$ \\
Scheduler type              & Cosine Annealing \\
Random seed                 & 42 \\
\bottomrule
\end{tabular}
\vspace{0.5em}
\label{tab:pretrain-hyperparams}
\end{table}

\textbf{Hardware and Training Regime:}
All training and evaluation were conducted on a single GPU using CSC's Puhti supercomputing environment \cite{puhti}. Training one model typically lasted 8--14 hours, depending on model depth and training phase.

\subsection{Environment and Libraries}
All model scripts were implemented using \textit{PyTorch} as the primary and only framework. Training and evaluation scripts were developed mainly using PyTorch modules, with support from standard libraries such as \textit{HuggingFace Tokenizers and Evaluation Datasets}, as well as Python standard libraries. For data loading, preprocessing, and tokenization, custom scripts were used, built around the standard PyTorch \texttt{Dataset} and \texttt{DataLoader} interfaces.

All code and configuration files related to this work were developed to support reproducibility and can be easily adapted for scaling up to larger models or for further modifications. They are available at GitHub \footnote{\url{https://github.com/Jtapsa/Decoder_Only_Transformer_Research}}.

\section{Training}

\subsection{Training Process}
The training was conducted in two phases to progressively validate the parallel path architecture and to test its flexibility from a training perspective.

In the initial phase, baseline models and their variants were trained using only the \textit{TinyStories} dataset to validate core functionality. This phase included LLaMA-based models with hidden dimensions of 256 and 192 (denoted as \texttt{LLaMA\_256} and \texttt{LLaMA\_192}, respectively), in addition to the parallel model variants.

Next, a second training phase was conducted using a \textit{combined dataset} consisting of TinyStories and 20\% of the NVIDIA/OpenMathInstruct-1 dataset. 

In this phase:

\begin{itemize}
    \item \texttt{LLaMA\_256} and \texttt{LLaMA\_192} were trained from scratch on the full combined dataset.
    \item Two smaller parallel-path models, \texttt{Path\_1} and \texttt{Path\_2} were pretrained independently:
    \begin{itemize}
        \item \texttt{Path\_1} was trained on 60\% of TinyStories.
        \item \texttt{Path\_2} was trained on 60\% of the 20\%  OpenMathInstruct-1 subset.
    \end{itemize}
\end{itemize}

After pretraining, these two smaller models (Path 1 and 2) were \textbf{combined into composite models} and further \textbf{pretrained jointly} using the remaining 40\% of the dataset.

\begin{figure}[ht]
\centering
\includegraphics[width=0.45\textwidth]{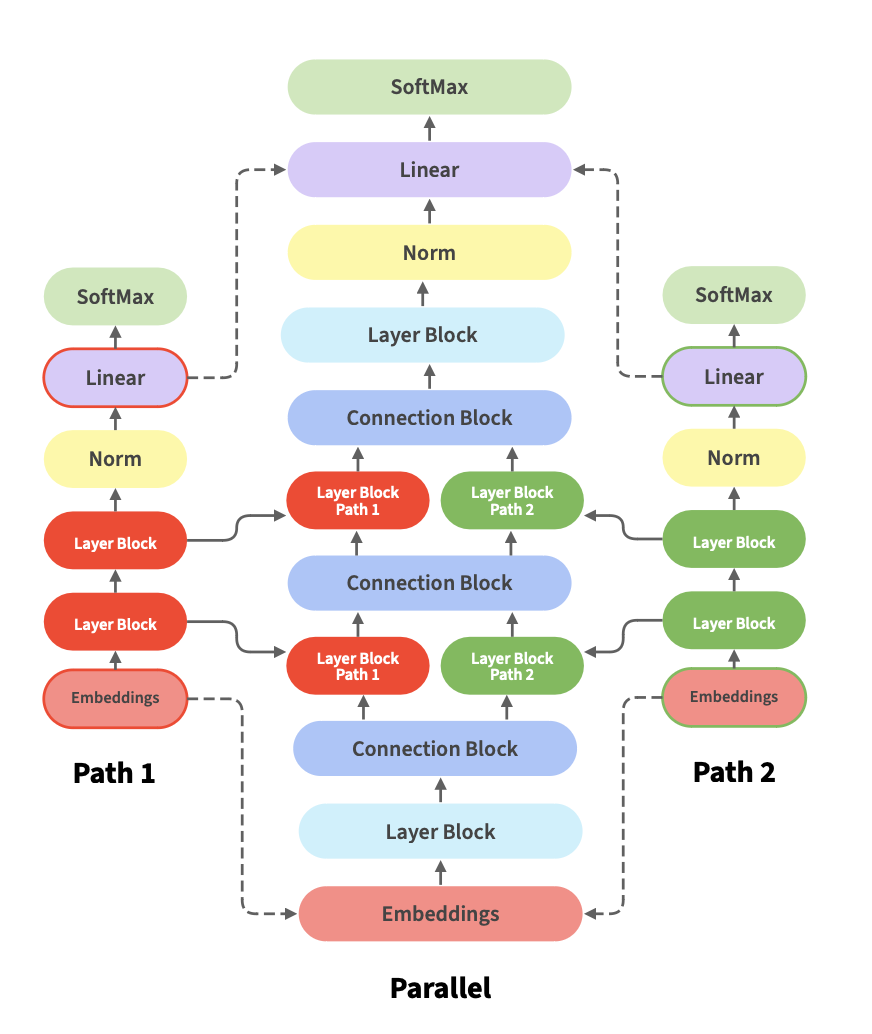}
\Description{..}
\caption[]{
Example of a constructed parallel model from two independently pretrained sub-paths. \texttt{Path\_1} and \texttt{Path\_2} are trained separately using different datasets. These pretrained components are merged to create a composite model with parallel layer blocks. Solid arrows indicate reused components (e.g., QKV and FFN weights), while dashed arrows represent the concatenation of embedding and final linear projection weights from \texttt{Path\_1} and \texttt{Path\_2}. \textbf{Note:} The figure shows a smaller-scale example for clarity.}
\label{fig:parallel-combination}
\end{figure}

This composite model integration followed the strategy illustrated in Figure~\ref{fig:parallel-combination}, where all Layer Block weights in the parallel subpaths is from pre-trained \texttt{Path\_1} and \texttt{Path\_2}. Additionally, the embedding and final linear projection layer weights were as well initialized by concatenating the corresponding weights from the pretrained paths. Since the hidden dimension of the parallel composite model matched the sum of the dimensions of the individual paths (e.g., 128 + 128 = 256), this concatenation was possible without the need for reshaping.

These staggered training phases enabled controlled architectural exploration where the functionality of the parallel architecture was validated before testing whether the parallel paths could be constructed from pre-trained paths trained on different types of data, and whether the model would subsequently utilize the corresponding path based on the specific input.

\subsection{Training Data}
The training process used the following datasets:

\begin{itemize}

    \item \textbf{TinyStories}: A synthetic corpus designed for pretraining SLMs and TLMs on story-based content (children’s stories) \footnote{\url{https://huggingface.co/datasets/roneneldan/TinyStories}}. It was used as the sole training dataset in the first phase and as part of the combined training data in the second phase.

    \item \textbf{NVIDIA/OpenMathInstruct-1 (20\% subset)}: A dataset focusing on math and instruction-following tasks \footnote{\url{https://huggingface.co/datasets/nvidia/OpenMathInstruct-1}}. The \textbf{question} and \textbf{generated\_solution} fields from the dataset were combined to form each training example. This work used a 20\% sample of the full dataset to complement \textbf{TinyStories} with data from a different domain (math and reasoning), aiming to test whether the parallel models naturally learn to utilize the appropriate domain-specific path—trained either on \textbf{TinyStories} or on the \textbf{NVIDIA/OpenMathInstruct-1}.
    
\end{itemize}

\subsection{Tokenizers}

Two tokenizers were used, corresponding to the phases of training:

\begin{itemize}

    \item In the \textbf{initial training phase}, the \texttt{TinyStories 3M} tokenizer \footnote{\url{https://huggingface.co/roneneldan/TinyStories-3M}} where used from HuggingFace.

    \item For the \textbf{second combined training phase}, the tokenizer was switched to the \texttt{EleutherAI/gpt-neox-20b} tokenizer \footnote{\url{https://huggingface.co/EleutherAI/gpt-neox-20b}}, which provides more suitable compatibility with OpenMathInstruct and aligns with the GPT-style vocabulary structure. 

\end{itemize}

\subsection{Pre-Tokenizing}

To ensure a fair comparison between models, all training examples were \textbf{pre-tokenized} into a contiguous token stream and then randomly split into fixed-length chunks of 256 tokens. Individual examples were separated using an end-of-sentence (EOS) token. This was done at the data preprocessing level to ensure that examples from different datasets were not mixed within a single training chunk. The splitting process was repeated twice so that both training epochs contained different, non-overlapping sequences.

Using this strategy, each training epoch contained a total of 477 million tokens from \textbf{TinyStories} and 206 million tokens from the \textbf{NVIDIA/OpenMathInstruct-1} subset.

In the second training phase, which involved both datasets, the pre-tokenized chunks from \textbf{TinyStories} and \textbf{OpenMathInstruct-1} were divided into two sub-collections: 60\% and 40\%. The base models were trained directly on the combined 100\% collection. \texttt{Path\_1} was trained on the 60\% sub-collection of \textbf{TinyStories}, while \texttt{Path\_2} was trained on the 60\% sub-collection of \textbf{OpenMathInstruct-1}. The parallel models—composite architectures that merged weights from \texttt{Path\_1} and \texttt{Path\_2}, were trained on the remaining 40\% sub-collections from both datasets.

During training, the chunks from sub-collections were shuffled so that training batches contained a mix of examples from both sources. This ensured that the base and parallel models were trained on the exact same chunks and total number of tokens, even though the composite parallel models used only the 40\% sub-collections. The missing 60\% sub-collections had already been used to pre-train \texttt{Path\_1} and \texttt{Path\_2}.

\section{Performance and Evaluation}

\subsection{Training Time Comparison}

Although models performance is the primary focus of this work, training time related efficiency provide important aspect into the possible benefits using of the proposed architecture. In the second training phase, where the combined TinyStories + OpenMathInstruct-1 dataset was used, expected result was confirmed that models, where parallel paths were used, required notably \textbf{less total training time} compared to their LLaMA counterparts as Table~\ref{tab:training-time} shows. Indeed, $\text{Parallel}^2$ demonstrates the case where paths were trained using an individual pipeline, resulting in the Stage 1 training time corresponding to the longest individual path training time.

This efficiency gain is due in part to the architectural \textbf{modularity} of the parallel-path design. Each subpath (e.g., \texttt{Path\_1} and \texttt{Path\_2}) uses smaller dimensionality, fewer layers and are trained fully \textbf{independently}. Those parallel paths can be trained also using own environments, which reduce training time evermore. 
Like this work shows, those individual paths can be fused into a composite model and further train to work jointly as parallel paths. This trait makes parallel path models more adaptable to limited compute environments, also allowing staggered or distributed training.

\begin{table}[ht]
\centering
\caption[]{Approximate training time (in hours) during the second training phase.}
\begin{tabular}{lccc}
\toprule
\textbf{Model} & \textbf{Stage 1 (h)} & \textbf{Stage 2 (h)} & \textbf{Total} \\
\midrule
LLaMA\_256  & -- & 12.5 & 12.5 \\
LLaMA\_192  & -- & 11 & 11.0 \\
Path\_1     & 3.5  & -- & 3.5 \\
Path\_2     & 1.5 & -- & 1.5 \\
Parallel    & (3.5 + 1.5) & 4.35 & \textbf{9.35} \\
$\text{Parallel}^2$   & 3.5 & 4.35 & \textbf{7.85} \\
\bottomrule
\end{tabular}
\vspace{0.5em}
\label{tab:training-time}
\end{table}

This and model evaluation result in \ref{sec:model_eval} demonstrate that the parallel path architecture is not only effective but also \textbf{training efficient} and well suitable for modular or distributed training pipelines.

\subsection{Model Evaluation}
\label{sec:model_eval}

Model evaluations closely followed the methodologies provided in the \texttt{\detokenize{babylm/evaluation-pipeline-2024}} \cite{babylm-2024}. However, not all tasks were included in the more lightweight evaluation pipeline, which this work followed. This pipeline applied the full BLiMP \cite{warstadt2020blimp} benchmark  and a selected subset of tasks from the GLUE \cite{wang2018glue} and SuperGLUE \cite{wang2019superglue} benchmarks. The GLUE evaluation covered following sub-tasks: \texttt{CoLA}, \texttt{SST-2}, \texttt{MRPC}, \texttt{STS-B}, \texttt{QQP}, \texttt{MNLI}, \texttt{QNLI}, and \texttt{RTE}. From SuperGLUE, \texttt{BoolQ}, \texttt{WSC}, and \texttt{MultiRC} were included in the evaluation pipeline. 

This combination of BLiMP, GLUE and SuperGLUE tasks cover sentiment analysis (SST-2), paraphrase decetion (MRPC, QQP), natural language inference and question answering (MNLI, QNLI, RTE), factual reasoning (BoolQ), linguistic acceptability (CoLA), pronoun resolution (WSC), multi-sentence understanding (MultiRC), all under low-resource fine-tuning.


\subsection{Fine-Tuning}

\subsubsection{LoRA Fine-Tuning on SuperGLUE and GLUE}

All models were fine-tuned using the \textbf{LoRA (Low-Rank Adaptation)} framework, which enable parameter-efficient adaptation to handle corresponding \textbf{GLUE} and \textbf{SuperGLUE} evaluation tasks. Evaluation followed the outline of the evaluation setup provided in the "babylm/evaluation-pipeline-2024"\footnote{\url{https://github.com/babylm/evaluation-pipeline-2024}}, with minor modifications (e.g., patience), applying the corresponding hyperparameters across all tasks and models. These hyperparameters (see Table~\ref{tab:training-hyperparams}) include standard optimization settings such as learning rate, weight decay, and a scheduler type as well as LoRA-specific parameters (rank, alpha, dropout).

\begin{table}[ht]
\centering
\caption[]{Training hyperparameters used across all experiments.}
\begin{tabular}{ll}
\toprule
\textbf{Hyperparameter}     & \textbf{Value} \\
\midrule
Optimizer                   & AdamW \\
Learning rate               & $1 \times 10^{-3}$ \\
Batch size                  & 64 \\
Maximum epochs              & 32 \\
Evaluate every (epochs)     & 1 \\
Patience                    & 5 \\
LoRa alpha                  & 16 \\
LoRA rank                   & 8 \\
LoRA dropout                & 0.1 \\
Sequence length             & 128 \\
Weight decay                & 0.1 \\
Scheduler type              & Linear \\
\bottomrule
\end{tabular}
\vspace{0.5em}
\label{tab:training-hyperparams}
\end{table}

A prominent observation related to model performance was, in some cases, a \textit{high sensitivity to the random seed} used for weight initialization during fine-tuning. 

In the worst case, the observed variability reached up to \textbf{12\%} in the corresponding evaluation metric across runs, due to initialization seed differences in the LoRA-injected linear layers. This issue notably occurred only for tasks with smaller fine-tuning datasets.

To address this problem, the final evaluation included \textbf{multiple fine-tuning runs (5 different seeds: 3, 12, 42, 100, 200)} for each model, and the reported results are \textbf{averaged scores} across runs. Exceptions were made for tasks such as MNLI (2 seeds: (12, 42)) and QQP (1 seed: 42), due to the size of the fine-tuning dataset, the required time for training and evaluation, and the low observed variability in scores across random seeds.

These findings underline the importance of seed-averaged evaluation, especially in case of small decoder-only models, which uses a fine-tuned additional classifier layer to perform GLUE and SuperGLUE evaluation tasks.

\subsection{Generation Capabilities}

\label{sec:gen_capalbilities}

In addition to benchmark evaluations, a qualitative manual evaluation was conducted to measure the models generation behavior. A set of narrative and instruction-style prompts was used to compare responses from different models. Even though the models' training data included math-instruction-related content, the following prompt analysis focuses solely on story-style (TinyStories) prompts, due to the models' inability to generate accurate or at times even reasonable continuations for the math-style prompt.

The models used in this qualitative evaluation were taken from the second training phase and include: \\

\texttt{LLaMA\_256},

{\ttfamily
Parallel\_Gumb\allowbreak\_MoE\allowbreak\_1},

{\ttfamily
Parallel\_Gumb\allowbreak\_MoE\allowbreak\_2} and

{\ttfamily
Parallel\_Share\allowbreak\_Linear} \\

as listed in Table~\ref{tab:combined-eval-group}.

\vspace{1em}
\noindent\textbf{Prompt\_1:} \textit{Once upon a time, there was a little}
\vspace{1em}

\noindent\textbf{LLaMA\_256:} \\
\textit{Once upon a time, there was a little girl named Lily. She loved to play outside in the sunshine...}

\textbf{Next token predictions:} girl — $100\%$ \\


\noindent\textbf{Parallel\_Gumb\_MoE\_1:} \\
\textit{Once upon a time, there was a little boy named Tim. Tim loved to play outside with his friends...}

\textbf{Next token predictions:} girl — $99.5\%$ \: boy — $0.5\%$ \\


\noindent\textbf{Parallel\_Gumb\_MoE\_2:} \\
\textit{Once upon a time, there was a little girl named Lily. She loved to play with her toys and run around outside...}

\textbf{Next token predictions:} girl — $99.5\%$ \: boy — $0.5\%$ \\


\noindent\textbf{Parallel\_Share\_Linear:} \\
\textit{Once upon a time, there was a little girl named Lily. She loved to play outside in the park with her friends...}

\textbf{Next token predictions:} girl — $100\%$ \\

 \vspace{1em}

\noindent The first prompt \textbf{'Once upon a time, there was a little'} tested general story generation capabilities, following the typical narrative opening style commonly found in the TinyStories training dataset. As observed, all models produced fluent, story-like continuations. However, \textbf{Parallel\_Gumb\_MoE\_1} and \textbf{Parallel\_Gumb\_MoE\_2} assigned a low probability to the main character being a boy. Notably, for the prompt continuations  involving a 'girl', all models consistently predicted the name to be Lily. In the case of \textbf{Parallel\_Gumb\_MoE\_1} and \textbf{Parallel\_Gumb\_MoE\_2}, when the story continued with "boy", the name "Timmy" was predicted with $100\%$ probability. 

\vspace{1em}

\noindent\textbf{Prompt\_2:} \textit{Once upon a time, there was a little boy named Lily.}
\vspace{1em}

\noindent\textbf{LLaMA\_256:} \\
\textit{Once upon a time, there was a little boy named Lily. She loved to play outside in the sun...}

\textbf{Next token predictions:} She — $99.85\%$ \: Lily — $0.12\%$ \\


\noindent\textbf{Parallel\_Gumb\_MoE\_1:} \\
\textit{Once upon a time, there was a little boy named Lily. She loved to play in the snow...}

\textbf{Next token predictions:} She — $99.98\%$ \: Lily — $0.02\%$ \\


\noindent\textbf{Parallel\_Gumb\_MoE\_2:} \\
\textit{Once upon a time, there was a little boy named Lily. She loved to play with her toys and run around outside...}

\textbf{Next token predictions:} She — $100\%$ \\


\noindent\textbf{Parallel\_Share\_Linear:} \\
\textit{Once upon a time, there was a little boy named Lily. He loved to play outside in the park...}

\textbf{Next token predictions:}  He — $95.5\%$ \: She — $5.5\%$ \\

\vspace{1em}

\noindent The second prompt, \textbf{"Once upon a time, there was a little boy named Lily."}, evaluated the model’s ability to interpret gender cues, specifically the term "boy", and to maintain consistency in gendered pronoun usage. All models except \textbf{Parallel\_Share\_Linear} incorrectly associated gender with the name rather than relying on the contextual clue provided by the noun "boy". However \textbf{Parallel\_Share\_Linear} led to a small probability being assigned to the incorrect pronoun ("She") instead of the appropriate one ("He").

\vspace{1em}

\noindent\textbf{Prompt\_3:} \textit{There was a little girl.}
\vspace{1em}

\noindent\textbf{LLaMA\_256:} \\
\textit{There was a little girl. She was only three years old...}

\textbf{Next token predictions:} She — $100\%$ \\


\noindent\textbf{Parallel\_Gumb\_MoE\_1:} \\
\textit{There was a little girl. She was so excited to play with her new pet...}

\textbf{Next token predictions:} She — $100\%$ \\


\noindent\textbf{Parallel\_Gumb\_MoE\_2:} \\
\textit{There was a little girl. She was very happy and she loved to play with her toys...}

\textbf{Next token predictions:} She — $100\%$ \\


\noindent\textbf{Parallel\_Share\_Linear:} \\
\textit{There was a little girl. She was very happy and she loved to play with her toys...}

\textbf{Next token predictions:} She — $100\%$ \\

\noindent The third prompt was \textbf{‘There was a little girl.’} In this case, the models were intentionally prompted with a sentence containing only a gender cue (“girl”), without any additional name-based information. All models correctly followed with the appropriate pronoun (“She”).

\vspace{1em}

\noindent\textbf{Prompt\_4:} \textit{There was a little girl named Timmy.}
\vspace{1em}

\noindent\textbf{LLaMA\_256:} \\
\textit{There was a little girl named Timmy. Timmy loved to play outside in the sun... He wanted to go ...}

\textbf{Next token predictions:} Tim (Sub-token of Timmy) — $100\%$ \\


\noindent\textbf{Parallel\_Gumb\_MoE\_1:} \\
\textit{There was a little girl named Timmy. Timmy loved to play with his toys and run around outside...}

\textbf{Next token predictions:} Tim (Sub-token of Timmy) — $94.92\%$  \: She — $5.08\%$ \\ 


\noindent\textbf{Parallel\_Gumb\_MoE\_2:} \\
\textit{There was a little girl named Timmy. Timmy loved to play with his toys and run around outside...}

\textbf{Next token predictions:} Tim (Sub-token of Timmy) — $99.90\%$  \: She — $0.1\%$ \\ 


\noindent\textbf{Parallel\_Share\_Linear:} \\
\textit{There was a little girl named Timmy. Timmy loved to play outside in the park... He ran to the dog ...}

\textbf{Next token predictions:} Tim (Sub-token of Timmy) — $87.44\%$  \: She — $12.56\%$ \\

\noindent The fourth prompt was \textbf{‘There was a little girl named Timmy.’} This prompt again evaluated the models’ ability to interpret gender cues and maintain consistency in gendered pronoun usage, but with a different structure and the inclusion of the term “girl.” When the continued token was “Tim,” all models incorrectly associated gender with the name rather than relying on the contextual cue provided by the noun “girl.” Among the parallel models, \textbf{Parallel\_Share\_Linear} performed best, assigning some probability to correctly continuing the story with the appropriate pronoun (“she”).

\subsubsection{Routing and Path Utilization}

\label{sec:routing_path_utilization}

To analyze path utilization in the parallel models, prompts strongly associated with either math/instruction-style content or narrative story-based content were used. Since \textbf{Path\_1} was trained solely on \textbf{TinyStories}, and \texttt{Path\_2} on \textbf{OpenMathInstruct-1}, the optimal behavior for the parallel models would be to utilize the corresponding domain-specific path during next-token prediction.

For the Gumbel MoE variants, routing behavior can be analyzed by examining the \texttt{argmax} of the routing distribution at each layer, indicating whether the model selected \texttt{Path\_1}, \texttt{Path\_2}, or a combined representation.

In the case of \textbf{Parallel\_Share\_Linear}, no strong evidence of explicit path selection is present, as the output is a fixed linear combination of the parallel paths. However, by comparing each parallel path’s output to the combined output using \textbf{cosine similarity}, it is possible to assess which path had a more dominant influence on the final output at each layer. 

\begin{table}[ht]
\centering
\caption[]{Path selection (by parallel layer) for each model and prompt. For the Gumbel MoE variants, the values indicate the selected path (1 or 2) or the use of a combined representation (3) as determined by the routing mechanism. For the \texttt{Parallel\_Share\_Linear} model, the values (1 or 2) indicate which path has the highest cosine similarity with the combined output of the parallel layer.} 
\begin{tabular}{lccc}
\toprule
\textbf{Prompt} & \makecell{Parallel\_Gumb\\MoE\_1} & \makecell{Parallel\_Gumb\\MoE\_2} & \makecell{Parallel\_Share\\Linear} \\
\midrule
\rowcolor{gray!10}
Prompt\_1 & \textcolor{red}{2} → \textcolor{red}{2} → \textcolor{ForestGreen}{1} & \textcolor{red}{2} → \textcolor{red}{2} → \textcolor{Dandelion}{3} & \textcolor{red}{2} → \textcolor{ForestGreen}{1} \\

Prompt\_2 & \textcolor{red}{2} → \textcolor{red}{2} → \textcolor{red}{2} & \textcolor{red}{2} → \textcolor{red}{2} → \textcolor{red}{2} & \textcolor{ForestGreen}{1} → \textcolor{ForestGreen}{1} \\

\rowcolor{gray!10}
Prompt\_3 & \textcolor{red}{2} → \textcolor{red}{2} → \textcolor{ForestGreen}{1} & \textcolor{ForestGreen}{1} → \textcolor{ForestGreen}{1} → \textcolor{red}{2} & \textcolor{ForestGreen}{1} → \textcolor{ForestGreen}{1} \\

Prompt\_4 & \textcolor{red}{2} → \textcolor{red}{2} → \textcolor{red}{2} & \textcolor{red}{2} → \textcolor{red}{2} → \textcolor{Dandelion}{3} & \textcolor{ForestGreen}{1} → \textcolor{ForestGreen}{1} \\

\rowcolor{gray!10}
'Little girl' & \textcolor{red}{2} → \textcolor{red}{2} → \textcolor{ForestGreen}{1} & \textcolor{red}{2} → \textcolor{red}{2} → \textcolor{red}{2} & \textcolor{red}{2} → \textcolor{ForestGreen}{1} \\

\makecell{'He liked to \\ play with \\ his toys'} & \textcolor{red}{2} → \textcolor{ForestGreen}{1} → \textcolor{ForestGreen}{1} & \textcolor{red}{2} → \textcolor{ForestGreen}{1} → \textcolor{ForestGreen}{1} & \textcolor{ForestGreen}{1} → \textcolor{red}{2} \\

\rowcolor{gray!10}
'One day' & \textcolor{red}{2} → \textcolor{ForestGreen}{1} → \textcolor{red}{2} & \textcolor{red}{2} → \textcolor{red}{2} → \textcolor{red}{2} & \textcolor{red}{2} → \textcolor{ForestGreen}{1} \\

\makecell{'wen to \\ the park'} & \textcolor{red}{2} → \textcolor{ForestGreen}{1} → \textcolor{red}{2} & \textcolor{red}{2} → \textcolor{ForestGreen}{1} → \textcolor{red}{2} & \textcolor{ForestGreen}{1} → \textcolor{ForestGreen}{1} \\

\rowcolor{gray!10}
'yummy soup' &  \textcolor{ForestGreen}{1} → \textcolor{ForestGreen}{1} → \textcolor{ForestGreen}{1} &  \textcolor{ForestGreen}{1} → \textcolor{red}{2} → \textcolor{red}{2} & \textcolor{red}{2} → \textcolor{ForestGreen}{1} \\

\makecell[l]{'She loved \\ playing with'} & \textcolor{red}{2} → \textcolor{ForestGreen}{1} → \textcolor{ForestGreen}{1} & \textcolor{red}{2} → \textcolor{ForestGreen}{1} → \textcolor{red}{2} & \textcolor{ForestGreen}{1} → \textcolor{red}{2} \\

\bottomrule

\rowcolor{gray!10}
'Solve this' & \textcolor{red}{1} → \textcolor{red}{1} → \textcolor{ForestGreen}{2} & \textcolor{red}{1} → \textcolor{ForestGreen}{2} → \textcolor{ForestGreen}{2} & \textcolor{ForestGreen}{2} → \textcolor{red}{1} \\

\makecell{' Express \\ your \\ answer'} & \textcolor{red}{1} → \textcolor{red}{1} → \textcolor{red}{1} & \textcolor{ForestGreen}{2} → \textcolor{ForestGreen}{2} → \textcolor{red}{1} & \textcolor{red}{1} → \textcolor{red}{1} \\

\rowcolor{gray!10}
'<llm-code>' & \textcolor{red}{1} → \textcolor{red}{1} → \textcolor{red}{1} & \textcolor{ForestGreen}{2} → \textcolor{red}{1} → \textcolor{ForestGreen}{2} & \textcolor{red}{1} → \textcolor{red}{1} \\

\makecell{'Find the \\ largest'} & \textcolor{red}{1} → \textcolor{red}{1} → \textcolor{ForestGreen}{2} & \textcolor{ForestGreen}{2} → \textcolor{ForestGreen}{2} → \textcolor{ForestGreen}{2} & \textcolor{ForestGreen}{2} → \textcolor{ForestGreen}{2} \\

\rowcolor{gray!10}
'Evaluate' & \textcolor{red}{1} → \textcolor{ForestGreen}{2} → \textcolor{ForestGreen}{2} & \textcolor{ForestGreen}{2} → \textcolor{ForestGreen}{2} → \textcolor{red}{1} & \textcolor{red}{1} → \textcolor{ForestGreen}{2} \\

\makecell{'from sympy \\ import sqrt'} & \textcolor{red}{1} → \textcolor{red}{1} → \textcolor{ForestGreen}{2} & \textcolor{ForestGreen}{2} → \textcolor{ForestGreen}{2} → \textcolor{Dandelion}{3} & \textcolor{ForestGreen}{2} → \textcolor{red}{1} \\

\rowcolor{gray!10}
'Simplify' & \textcolor{red}{1} → \textcolor{ForestGreen}{2} → \textcolor{ForestGreen}{2} & \textcolor{ForestGreen}{2} → \textcolor{ForestGreen}{2} → \textcolor{ForestGreen}{2} & \textcolor{red}{1} → \textcolor{red}{1} \\

\makecell{'Let's solve \\ this problem'} & \textcolor{red}{1} → \textcolor{Dandelion}{3} → \textcolor{ForestGreen}{2} & \textcolor{Dandelion}{3} → \textcolor{red}{1} → \textcolor{red}{1} & \textcolor{red}{1} → \textcolor{red}{1} \\

\rowcolor{gray!10}
'Python code' &  \textcolor{ForestGreen}{2} → \textcolor{red}{1} → \textcolor{red}{1} &  \textcolor{ForestGreen}{2} → \textcolor{ForestGreen}{2} → \textcolor{red}{1}  &  \textcolor{red}{1} → \textcolor{red}{1}  \\

'1 + 1' &  \textcolor{ForestGreen}{2} → \textcolor{ForestGreen}{2} → \textcolor{ForestGreen}{2} &  \textcolor{ForestGreen}{2} → \textcolor{ForestGreen}{2} → \textcolor{ForestGreen}{2} & \textcolor{ForestGreen}{2} → \textcolor{red}{1} \\


\bottomrule

\makecell{\textbf{Path\_1}} & 48\% & 23\% & 70\% \\
\makecell{\textbf{Path\_2}} & 50\% & 70\% & 30\% \\
\makecell{\textbf{Combined}} & 2\% & 7\% & -- \\
\bottomrule
\makecell{\textbf{Accuracy}} & 40\% & 46.5\% & 50 \\
\bottomrule

\end{tabular}
\vspace{0.5em}
\label{tab:path_utilization}
\end{table}

Table~\ref{tab:path_utilization} exhibits an interesting perspective of the selection and utilization of parallel-model path. Note that \textit{Parallel\_Share\_Linear} includes only two parallel layers, as the final layer expands the output dimension to match that of the \texttt{Layer Block}. Green and red colors indicate whether the path selection was correct or incorrect based on the prompt type, while yellow highlights cases where the Gumbel MoE variants selected a combined representation instead of either individual path. For models only \textbf{Parallel\_Gumb\_MoE\_1} have evenly distributed path selection, which would be natural, because of 'correct' paths is evenly distributed. Despite of that \textbf{Parallel\_Gumb\_MoE\_1} have weakest accuracy in table~\ref{tab:path_utilization} correct path selection measurement. This indicates that if model uses more dominantly Path\_1 or Path\_2, this directly correlates to measurement accuracy. Interestingly, models evaluation result as well follows this insight, which can be seen from table~\ref{tab:combined-eval-group}.

\section{Discussion} 

\subsection{Evaluation Results}

The evaluation results show that the Gumbel MoE model variants did not achieve performance gains compared to the baseline models. Among them, only \textbf{Parallel\_Gumb\_MoE\_2} was able to outperform the smaller baseline \textbf{LLaMA\_192}, and it approached the performance of the larger \textbf{LLaMA\_256} model, when the parallel paths were pre-trained. However, the other Gumbel MoE variant, \textbf{Parallel\_Gumb\_MoE\_1}, underperformed despite exhibiting a more balanced path selection behavior, as shown in Table~\ref{tab:path_utilization}.

This suggests a potential limitation in the current Gumbel MoE routing mechanism and structure, which fail to specialize each path effectively and potentially led to degraded task performance. 

In contrast, the more lightweight \textbf{Parallel\_Share\_Linear} model, which simply combines parallel path outputs using a linear projection layer, achieved the highest overall results in both training phases. It tied with \textbf{LLaMA\_256} during the first phase, where all models were trained from scratch, and outperformed all other models during the second phase, where the parallel paths were pre-trained independently.

These results indicate that parallel architectures can yield performance benefits in addition to training efficiency, but only when the paths are connected using an appropriately effective mechanism.

However, the current setup still leaves room for improvement in all parallel models. Performance may be further enhanced through architectural supplements such as residual connections, normalization strategies and alternative auxiliary loss functions in the Gumbel MoE variants.

\begin{table}[ht]
\centering
\caption[]{Evaluation results for models trained on TinyStories only and on TinyStories + OpenMathInstruct.}
\begin{tabular}{lcccc}
\toprule
\textbf{Model} & \textbf{BLiMP ↑} & \textbf{GLUE ↑} & \makecell{\textbf{Super} \\ \textbf{GLUE ↑}} & \makecell{\textbf{Macro-} \\ \textbf{average ↑}} \\
\midrule
\multicolumn{5}{c}{\textbf{TinyStories only}} \\
\midrule
LLaMA\_256                           & 64.30            & \textbf{59.40}   & 55.90           & \textbf{59.85} \\
\midrule
LLaMA\_192                           & \textbf{64.75}   & 59.10            & 55.20           & 59.70 \\
\midrule
\makecell{Parallel \\ Gumb\_MoE\_1}  & 62.30            & 57.65            & 54.50           & 58.15 \\
\midrule
\makecell{Parallel \\ Gumb\_MoE\_2}  & 61.85            & 58.80            & 54.75           & 58.45 \\
\midrule
\makecell{Parallel \\ Share\_Linear} & 64.15            & 58.65            & \textbf{56.70}  & \textbf{59.85} \\
\midrule
\multicolumn{5}{c}{\textbf{TinyStories + OpenMathInstruct}} \\
\midrule
LLaMA\_256                           & 62.55            & \textbf{59.85}   & 55.30           & 59.25 \\
\midrule
LLaMA\_192                           & 61.90            & 58.95            & 54.70           & 58.50 \\
\midrule
Path\_1                              & 60.80            & 56.95            & 52.95           & 56.90 \\
\midrule
Path\_2                              & 57.80            & 54.25            & 53.90           & 55.30 \\
\midrule
\makecell{Parallel \\ Gumb\_MoE\_1}  & 62.75            & 57.35            & 54.70           & 58.25 \\
\midrule
\makecell{Parallel \\ Gumb\_MoE\_2}  & \textbf{63.60}   & 57.95            & 55.10           & 58.90 \\
\midrule
\makecell{Parallel \\ Share\_Linear} & 63.20            & 58.60            & \textbf{56.40}  & \textbf{59.40} \\
\bottomrule
\end{tabular}
\label{tab:combined-eval-group}
\end{table}

\subsection{Limitations}

This work was conducted under limited computational resources, resulting in relatively small model sizes and reduced-scale training datasets. While the proposed architecture is designed to scale, its behavior has not yet been validated at the large language model (LLM) scale. In addition, the evaluation was limited to a small subset of the SuperGLUE, GLUE, and BLiMP benchmarks, which are better suited for assessing small models. The parallel path routing mechanism—based on Gumbel-Softmax—was not compared against other routing alternatives such as standard softmax and other methods. Furthermore, routing did not rely on additional expert networks, instead of that, combined linear layers and parallel paths themselves acted as functional equivalents to experts, minimizing parameter overhead.

\subsection{Architectural Insights}

The modularity of the parallel-path architecture provides several practical benefits. Most notably, the weights used in each parallel path can be \textbf{trained independently} and later recombined, significantly reducing total training time and enabling distributed or asynchronous pretraining workflows prior to model integration. In addition, the parallel-path architecture offers high flexibility for model customization and introduces a novel approach to information sharing across parallel paths, as well as sparse parameter activation.

\subsection{Future Work}

Several directions exist for extending this work:
\begin{itemize}
    \item \textbf{Specialized Parallel Paths}: Future models could use customized paths, one with standard attention, another with different, to specialize for different linguistic features or task types.
    \textbf{different dimensionalities} across different paths could be explored, enabling more diverse sub-network architectures.
    \item \textbf{More than Two Paths}: Expanding to several parallel paths could increase the model's representational diversity.
    \item \textbf{Different routing mechanisms}: Alternative routing strategies could enable better utilization of parallel paths.    
    \item \textbf{Larger-Scale Validation}: Future work will explore the architecture at higher parameter counts and on more extensive datasets (e.g., The Pile, OpenWebText2, or Multilingual C4).
\end{itemize}

\section{Conclusion}
In summary, this work explored the parallel paths integration methods in decoder-only transformer models using both linear combination and MoE-style routing mechanisms in addition to parameter reuse and flexible scaling. While the Gumbel MoE variants demonstrated the cautious potential of expert-style selection in path level, their performance was limited by comparison to dense base models. In contrast, the simpler \textbf{Parallel\_Share\_Linear} model not only matched or outperformed larger dense baselines but also demonstrated more reliable alignment between path utilization and input type. These findings suggest that modular parallel architectures paired with an appropriate integration mechanisms can improve model efficiency and performance. 

Future work should explore supplement routing structures, parallel layer level  normalization strategies with residual connections to fully explore the benefits of parallel model design.

\begin{acks}
This work was possible through CSC computation environments offered via University of Oulu ICT services.  I would like to thank Msc Moinul Islam for sharing helpful tips that contributed to the development of this research. This work is partly supported by Finnish Research Council Profi7 Hybrid Intelligence programme. 
\end{acks}


\bibliographystyle{ACM-Reference-Format}
\bibliography{references.bib}


\begin{thebibliography}{20}


\ifx \showCODEN    \undefined \def \showCODEN     #1{\unskip}     \fi
\ifx \showDOI      \undefined \def \showDOI       #1{#1}\fi
\ifx \showISBNx    \undefined \def \showISBNx     #1{\unskip}     \fi
\ifx \showISBNxiii \undefined \def \showISBNxiii  #1{\unskip}     \fi
\ifx \showISSN     \undefined \def \showISSN      #1{\unskip}     \fi
\ifx \showLCCN     \undefined \def \showLCCN      #1{\unskip}     \fi
\ifx \shownote     \undefined \def \shownote      #1{#1}          \fi
\ifx \showarticletitle \undefined \def \showarticletitle #1{#1}   \fi
\ifx \showURL      \undefined \def \showURL       {\relax}        \fi
\providecommand\bibfield[2]{#2}
\providecommand\bibinfo[2]{#2}
\providecommand\natexlab[1]{#1}
\providecommand\showeprint[2][]{arXiv:#2}

\bibitem[Brown et~al\mbox{.}(2020)]%
        {brown2020language}
\bibfield{author}{\bibinfo{person}{Tom Brown}, \bibinfo{person}{Benjamin Mann}, \bibinfo{person}{Nick Ryder}, \bibinfo{person}{Melanie Subbiah}, \bibinfo{person}{Jared Kaplan}, \bibinfo{person}{Prafulla Dhariwal}, \bibinfo{person}{Arvind Neelakantan}, \bibinfo{person}{Pranav Shyam}, \bibinfo{person}{Girish Sastry}, \bibinfo{person}{Amanda Askell}, {et~al\mbox{.}}} \bibinfo{year}{2020}\natexlab{}.
\newblock \showarticletitle{Language Models are Few-Shot Learners}.
\newblock \bibinfo{journal}{\emph{Advances in Neural Information Processing Systems}}  \bibinfo{volume}{33} (\bibinfo{year}{2020}), \bibinfo{pages}{1877--1901}.
\newblock


\bibitem[Choshen et~al\mbox{.}(2024)]%
        {babylm-2024}
\bibfield{author}{\bibinfo{person}{Leshem Choshen}, \bibinfo{person}{Ryan Cotterell}, \bibinfo{person}{Michael~Y. Hu}, \bibinfo{person}{Tal Linzen}, \bibinfo{person}{Aaron Mueller}, \bibinfo{person}{Candace Ross}, \bibinfo{person}{Alex Warstadt}, \bibinfo{person}{Ethan Wilcox}, \bibinfo{person}{Adina Williams}, {and} \bibinfo{person}{Chengxu Zhuang}.} \bibinfo{year}{2024}\natexlab{}.
\newblock \showarticletitle{[Call for Papers] The 2nd {BabyLM} {C}hallenge: Sample-efficient pretraining on a developmentally plausible corpus}.
\newblock \bibinfo{journal}{\emph{Computing Research Repository}}  \bibinfo{volume}{arXiv:2404.06214} (\bibinfo{year}{2024}).
\newblock
\urldef\tempurl%
\url{https://arxiv.org/abs/2404.06214}
\showURL{%
\tempurl}


\bibitem[Chowdhery et~al\mbox{.}(2022)]%
        {chowdhery2022palm}
\bibfield{author}{\bibinfo{person}{Aakanksha Chowdhery}, \bibinfo{person}{Sharan Narang}, \bibinfo{person}{Jacob Devlin}, {and} \bibinfo{person}{et al.}} \bibinfo{year}{2022}\natexlab{}.
\newblock \bibinfo{title}{PaLM: Scaling Language Modeling with Pathways}.
\newblock
\newblock
\showeprint[arxiv]{2204.02311}~[cs.CL]
\urldef\tempurl%
\url{https://arxiv.org/abs/2204.02311}
\showURL{%
\tempurl}


\bibitem[{CSC -- IT Center for Science}({[n.\,d.]})]%
        {puhti}
\bibfield{author}{\bibinfo{person}{{CSC -- IT Center for Science}}.} \bibinfo{year}{[n.\,d.]}\natexlab{}.
\newblock \bibinfo{title}{Puhti Supercomputer}.
\newblock \bibinfo{howpublished}{\url{https://docs.csc.fi/computing/systems-puhti/}}.
\newblock
\newblock
\shownote{Accessed: 2025-05-06}.


\bibitem[Du et~al\mbox{.}(2022)]%
        {du2022glam}
\bibfield{author}{\bibinfo{person}{Nan Du}, \bibinfo{person}{Le Hou}, \bibinfo{person}{Aitor Zhang}, \bibinfo{person}{Anton Bakhtin}, \bibinfo{person}{Nathan Scales}, \bibinfo{person}{Zhifeng Dai}, \bibinfo{person}{Xin Li}, \bibinfo{person}{Shixiang Xie}, \bibinfo{person}{William Fedus}, \bibinfo{person}{Mostafa Dehghani}, {and} \bibinfo{person}{et al.}} \bibinfo{year}{2022}\natexlab{}.
\newblock \showarticletitle{GLaM: Efficient Scaling of Language Models with Mixture-of-Experts}. In \bibinfo{booktitle}{\emph{International Conference on Learning Representations (ICLR)}}.
\newblock
\urldef\tempurl%
\url{https://openreview.net/forum?id=k7K6kB9td9}
\showURL{%
\tempurl}


\bibitem[Eldan and Li(2023)]%
        {eldan2023tinystoriessmalllanguagemodels}
\bibfield{author}{\bibinfo{person}{Ronen Eldan} {and} \bibinfo{person}{Yuanzhi Li}.} \bibinfo{year}{2023}\natexlab{}.
\newblock \bibinfo{title}{TinyStories: How Small Can Language Models Be and Still Speak Coherent English?}
\newblock
\newblock
\showeprint[arxiv]{2305.07759}~[cs.CL]
\urldef\tempurl%
\url{https://arxiv.org/abs/2305.07759}
\showURL{%
\tempurl}


\bibitem[Fedus et~al\mbox{.}(2021)]%
        {fedus2021switch}
\bibfield{author}{\bibinfo{person}{William Fedus}, \bibinfo{person}{Barret Zoph}, {and} \bibinfo{person}{Noam Shazeer}.} \bibinfo{year}{2021}\natexlab{}.
\newblock \showarticletitle{Switch Transformers: Scaling to Trillion Parameter Models with Simple and Efficient Sparsity}. In \bibinfo{booktitle}{\emph{Advances in Neural Information Processing Systems (NeurIPS)}}, Vol.~\bibinfo{volume}{34}. \bibinfo{pages}{8473--8483}.
\newblock
\urldef\tempurl%
\url{https://proceedings.neurips.cc/paper/2021/hash/2c5dc10619a37b0c79ef595e0bda0592-Abstract.html}
\showURL{%
\tempurl}


\bibitem[Fedus et~al\mbox{.}(2022)]%
        {fedus2022switchtransformersscalingtrillion}
\bibfield{author}{\bibinfo{person}{William Fedus}, \bibinfo{person}{Barret Zoph}, {and} \bibinfo{person}{Noam Shazeer}.} \bibinfo{year}{2022}\natexlab{}.
\newblock \bibinfo{title}{Switch Transformers: Scaling to Trillion Parameter Models with Simple and Efficient Sparsity}.
\newblock
\newblock
\showeprint[arxiv]{2101.03961}~[cs.LG]
\urldef\tempurl%
\url{https://arxiv.org/abs/2101.03961}
\showURL{%
\tempurl}


\bibitem[Jang et~al\mbox{.}(2017)]%
        {jang2017categorical}
\bibfield{author}{\bibinfo{person}{Eric Jang}, \bibinfo{person}{Shixiang Gu}, {and} \bibinfo{person}{Ben Poole}.} \bibinfo{year}{2017}\natexlab{}.
\newblock \showarticletitle{Categorical reparameterization with gumbel-softmax}. In \bibinfo{booktitle}{\emph{Proceedings of the 31st International Conference on Neural Information Processing Systems (NeurIPS)}}. \bibinfo{pages}{4107--4116}.
\newblock
\urldef\tempurl%
\url{https://proceedings.neurips.cc/paper/2017/file/7a98af17e63a0ac09ce2e96d03992fbc-Paper.pdf}
\showURL{%
\tempurl}


\bibitem[Lepikhin et~al\mbox{.}(2020)]%
        {lepikhin2020gshardscalinggiantmodels}
\bibfield{author}{\bibinfo{person}{Dmitry Lepikhin}, \bibinfo{person}{HyoukJoong Lee}, \bibinfo{person}{Yuanzhong Xu}, \bibinfo{person}{Dehao Chen}, \bibinfo{person}{Orhan Firat}, \bibinfo{person}{Yanping Huang}, \bibinfo{person}{Maxim Krikun}, \bibinfo{person}{Noam Shazeer}, {and} \bibinfo{person}{Zhifeng Chen}.} \bibinfo{year}{2020}\natexlab{}.
\newblock \bibinfo{title}{GShard: Scaling Giant Models with Conditional Computation and Automatic Sharding}.
\newblock
\newblock
\showeprint[arxiv]{2006.16668}~[cs.CL]
\urldef\tempurl%
\url{https://arxiv.org/abs/2006.16668}
\showURL{%
\tempurl}


\bibitem[Penedo et~al\mbox{.}(2023)]%
        {penedo2023falcon}
\bibfield{author}{\bibinfo{person}{Guillem Penedo}, \bibinfo{person}{Alexandre Almazouzi}, \bibinfo{person}{Inigo Jauregi~Unanue}, \bibinfo{person}{Ignacio Iacobacci}, \bibinfo{person}{Brian Davis}, \bibinfo{person}{Thomas Simeon}, \bibinfo{person}{Andrea Corda}, \bibinfo{person}{Guillaume Staerman}, {et~al\mbox{.}}} \bibinfo{year}{2023}\natexlab{}.
\newblock \bibinfo{title}{The RefinedWeb Dataset for Falcon LLMs: Outperforming Curated Corpora with Web Data, and the New Falcon Suite}.
\newblock
\newblock
\showeprint[arxiv]{2306.01116}~[cs.CL]


\bibitem[Peng et~al\mbox{.}(2022)]%
        {peng2022branchformerparallelmlpattentionarchitectures}
\bibfield{author}{\bibinfo{person}{Yifan Peng}, \bibinfo{person}{Siddharth Dalmia}, \bibinfo{person}{Ian Lane}, {and} \bibinfo{person}{Shinji Watanabe}.} \bibinfo{year}{2022}\natexlab{}.
\newblock \bibinfo{title}{Branchformer: Parallel MLP-Attention Architectures to Capture Local and Global Context for Speech Recognition and Understanding}.
\newblock
\newblock
\showeprint[arxiv]{2207.02971}~[cs.CL]
\urldef\tempurl%
\url{https://arxiv.org/abs/2207.02971}
\showURL{%
\tempurl}


\bibitem[Radford and Narasimhan(2018)]%
        {Radford2018ImprovingLU}
\bibfield{author}{\bibinfo{person}{Alec Radford} {and} \bibinfo{person}{Karthik Narasimhan}.} \bibinfo{year}{2018}\natexlab{}.
\newblock \showarticletitle{Improving Language Understanding by Generative Pre-Training}.
\newblock
\urldef\tempurl%
\url{https://api.semanticscholar.org/CorpusID:49313245}
\showURL{%
\tempurl}


\bibitem[Shazeer et~al\mbox{.}(2017)]%
        {shazeer2017outrageously}
\bibfield{author}{\bibinfo{person}{Noam Shazeer}, \bibinfo{person}{Azalia Mirhoseini}, \bibinfo{person}{Andrew Maziarz}, \bibinfo{person}{Krzysztof Davis}, \bibinfo{person}{Quoc Le}, \bibinfo{person}{Geoffrey Hinton}, {and} \bibinfo{person}{Jeff Dean}.} \bibinfo{year}{2017}\natexlab{}.
\newblock \showarticletitle{Outrageously large neural networks: The sparsely-gated mixture-of-experts layer}. In \bibinfo{booktitle}{\emph{International Conference on Learning Representations (ICLR)}}.
\newblock
\urldef\tempurl%
\url{https://openreview.net/forum?id=B1ckMDqlg}
\showURL{%
\tempurl}


\bibitem[Team(2024)]%
        {deepseek2024deepseekv2}
\bibfield{author}{\bibinfo{person}{DeepSeek Team}.} \bibinfo{year}{2024}\natexlab{}.
\newblock \bibinfo{title}{DeepSeek V2: Scaling Vision-Language Models with Mixture of Experts}.
\newblock
\newblock
\showeprint[arxiv]{2401.00733}~[cs.CL]


\bibitem[Touvron et~al\mbox{.}(2023)]%
        {touvron2023llama}
\bibfield{author}{\bibinfo{person}{Hugo Touvron}, \bibinfo{person}{Thibaut Lavril}, \bibinfo{person}{Gautier Izacard}, \bibinfo{person}{Xavier Martinet}, \bibinfo{person}{Marie-Anne Lachaux}, \bibinfo{person}{Timothée Lacroix}, \bibinfo{person}{Baptiste Rozière}, \bibinfo{person}{Naman Goyal}, \bibinfo{person}{Eric Hambro}, \bibinfo{person}{Faisal Azhar}, \bibinfo{person}{Aurelien Rodriguez}, \bibinfo{person}{Armand Joulin}, \bibinfo{person}{Edouard Grave}, {and} \bibinfo{person}{Guillaume Lample}.} \bibinfo{year}{2023}\natexlab{}.
\newblock \bibinfo{title}{LLaMA: Open and Efficient Foundation Language Models}.
\newblock
\newblock
\showeprint[arxiv]{2302.13971}~[cs.CL]
\urldef\tempurl%
\url{https://arxiv.org/abs/2302.13971}
\showURL{%
\tempurl}


\bibitem[Vaswani et~al\mbox{.}(2023)]%
        {vaswani2023attentionneed}
\bibfield{author}{\bibinfo{person}{Ashish Vaswani}, \bibinfo{person}{Noam Shazeer}, \bibinfo{person}{Niki Parmar}, \bibinfo{person}{Jakob Uszkoreit}, \bibinfo{person}{Llion Jones}, \bibinfo{person}{Aidan~N. Gomez}, \bibinfo{person}{Lukasz Kaiser}, {and} \bibinfo{person}{Illia Polosukhin}.} \bibinfo{year}{2023}\natexlab{}.
\newblock \bibinfo{title}{Attention Is All You Need}.
\newblock
\newblock
\showeprint[arxiv]{1706.03762}~[cs.CL]
\urldef\tempurl%
\url{https://arxiv.org/abs/1706.03762}
\showURL{%
\tempurl}


\bibitem[Wang et~al\mbox{.}(2019)]%
        {wang2019superglue}
\bibfield{author}{\bibinfo{person}{Alex Wang}, \bibinfo{person}{Yada Pruksachatkun}, \bibinfo{person}{Nikita Nangia}, \bibinfo{person}{Amanpreet Singh}, \bibinfo{person}{Julian Michael}, \bibinfo{person}{Felix Hill}, \bibinfo{person}{Omer Levy}, {and} \bibinfo{person}{Samuel~R Bowman}.} \bibinfo{year}{2019}\natexlab{}.
\newblock \showarticletitle{SuperGLUE: A Stickier Benchmark for General-Purpose Language Understanding Systems}. In \bibinfo{booktitle}{\emph{Advances in Neural Information Processing Systems}}, Vol.~\bibinfo{volume}{32}.
\newblock


\bibitem[Wang et~al\mbox{.}(2018)]%
        {wang2018glue}
\bibfield{author}{\bibinfo{person}{Alex Wang}, \bibinfo{person}{Amanpreet Singh}, \bibinfo{person}{Julian Michael}, \bibinfo{person}{Felix Hill}, \bibinfo{person}{Omer Levy}, {and} \bibinfo{person}{Samuel~R Bowman}.} \bibinfo{year}{2018}\natexlab{}.
\newblock \showarticletitle{GLUE: A Multi-Task Benchmark and Analysis Platform for Natural Language Understanding}. In \bibinfo{booktitle}{\emph{Proceedings of the 2018 EMNLP Workshop BlackboxNLP: Analyzing and Interpreting Neural Networks for NLP}}. \bibinfo{pages}{353--355}.
\newblock


\bibitem[Warstadt et~al\mbox{.}(2020)]%
        {warstadt2020blimp}
\bibfield{author}{\bibinfo{person}{Alex Warstadt}, \bibinfo{person}{Yining Cao}, \bibinfo{person}{Jun Ho}, \bibinfo{person}{Ellie Pavlick}, {and} \bibinfo{person}{Samuel~R Bowman}.} \bibinfo{year}{2020}\natexlab{}.
\newblock \showarticletitle{BLiMP: The Benchmark of Linguistic Minimal Pairs for English}.
\newblock \bibinfo{journal}{\emph{Transactions of the Association for Computational Linguistics}}  \bibinfo{volume}{8} (\bibinfo{year}{2020}), \bibinfo{pages}{377--392}.
\newblock


\end{thebibliography}

\end{document}